\documentclass[twoside]{article}
\usepackage{aistats2e}

\usepackage[sectionbib,round]{natbib}

\newcommand{\specialcell}[2][c]{\begin{tabular}[#1]{@{}c@{}}#2\end{tabular}}

\usepackage{multirow}
\usepackage{rotating}
\usepackage{subfigure}
\usepackage{tabularx}

\usepackage{mydefs}
\usepackage{mrwtd}

\begin{document}

\twocolumn[
\aistatstitle{Multi-relational Learning Using \\ Weighted Tensor Decomposition with Modular Loss}
\aistatsauthor{Ben London, Theodoros Rekatsinas, Bert Huang, and Lise Getoor}
\newcommand{\UMDAddress}{University of Maryland \\ College Park, MD 20742, USA}
\aistatsaddress{\UMDAddress \\ \texttt{\{blondon,thodrek,bert,getoor\}@cs.umd.edu}}
\runningtitle{Multi-relational Learning Using Weighted Tensor Decomposition}
\runningauthor{London, Rekatsinas, Huang, Getoor}
]

\begin{abstract}%
We propose a modular framework for multi-relational learning via tensor decomposition. In our learning setting, the training data contains multiple types of relationships among a set of objects, which we represent by a sparse three-mode tensor. The goal is to predict the values of the missing entries. To do so, we model each relationship as a function of a linear combination of latent factors. We learn this latent representation by computing a low-rank tensor decomposition, using quasi-Newton optimization of a weighted objective function. Sparsity in the observed data is captured by the weighted objective, leading to improved accuracy when training data is limited. Exploiting sparsity also improves efficiency, potentially up to an order of magnitude over unweighted approaches. In addition, our framework accommodates arbitrary combinations of smooth, task-specific loss functions, making it better suited for learning different types of relations. For the typical cases of real-valued functions and binary relations, we propose several loss functions and derive the associated parameter gradients. We evaluate our method on synthetic and real data, showing significant improvements in both accuracy and scalability over related factorization techniques.
\end{abstract}

\section{Introduction}
\label{sec:intro}

In network or relational data, one often finds multiple types of relations on a set of objects. For instance, in social networks, relationships between individuals may be personal, familial, or professional. We refer to this type of data as \define{multi-relational}.
In this paper, we propose a tensor decomposition model for transduction on multi-relational data. We consider a scenario in which we are given a fixed set of objects, a set of relations and a small training set, sampled from the full set of all potential pairwise relationships; our goal is to predict the unobserved relationships. The relations we consider may be binary-, discrete ordinal- or real-valued functions of the object pairs; for the binary-valued relationships, the training labels include both positive and negative examples.

There has been a growing interest in tensor methods within machine learning, partially due to their natural representation of multi-relational data \citep{kashima:sdm09}. Many contributions \citep{dunlavy:sandia06,dunlavy:tkdd11,gao:cidm11,xiong:sdm10} use the \define{canonical polyadic} (CP) decomposition, a generalization of singular value decomposition to tensors. Others \citep{bader:icdm07} have proposed models based on \define{decomposition into directional components} (DEDICOM) \citep{harshman:dedicom78}. We propose a similar decomposition (based on \citep{nickel:icml11}) which is more appropriate for multi-relational data, for reasons discussed in \autoref{sec:related_work}. Unlike these previous methods, we do not attempt to decompose the input tensor directly; rather, we explicitly model a mapping from the low-rank representation to the observed tensor, which is often better suited for prediction. For example, a binary relationship can be modeled as the sign of a latent representation; this gives the latent representation more freedom to increase the prediction margin, rather than reproduce $\TwoClass$ exactly.
In this respect, approaches like \define{maximum-margin matrix factorization} (MMMF) \citep{srebro:nips05a,rennie:icml05} and DEDICOM can be viewed as specializations of our framework.

\begin{figure*}[tp]
\begin{center}
\includegraphics[scale=0.5]{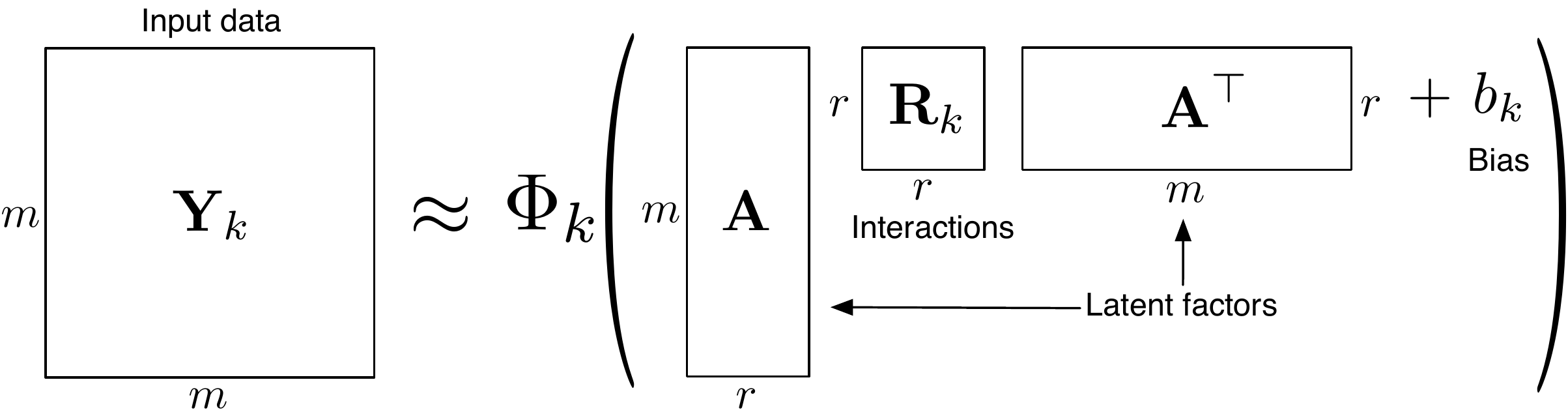}
\caption{For $k = 1,\dots,\nrel$, each slice $\Y_k$ of the input tensor is approximated by a function $\trans_k$ of a low-rank decomposition $\A \R_k \A\T + \bias_k$. The latent factors $\A$ are common to all slices. Each $\R_k$ determines the interactions of $\A$ in the $k\nth$ relation, while $\bias_k$ accounts for distributional bias.}
\label{fig:decomp}
\end{center}
\end{figure*}

Our proposed method, \define{\Algorithm} (see \autoref{sec:method}), assumes that the latent representation is determined by a linear combination of latent factors associated with each object. Learning these latent factors and their interactions in each relation thus becomes analogous to a weighted tensor decomposition (described in \autoref{sec:model} and illustrated in \autoref{fig:decomp}). We formulate this decomposition as a nonlinear optimization problem (\autoref{sec:objective}), which may incorporate any combination of smooth, task-specific loss functions. These task-specific loss functions allow simultaneous learning of various relation types, such as binary- and continuous-valued. By weighting the objective function, we are able to learn from limited observed (training) relationships without fitting the unobserved (testing) ones, improving both accuracy and efficiency.
We demonstrate the effectiveness of our approach in \autoref{sec:experiments}, using both real and synthetic data experiments.
Our results indicate that our approach is both more accurate and efficient than competing factorizations when training data is sparse.

\section{Preliminaries}
\label{sec:prelims}
This section introduces our notation and defines the problem of multi-relational transduction.

We denote tensors and matrices using bold, uppercase letters; similarly, we use bold, lowercase letters to denote vectors. For a tensor $\X$, let $\x_{i,j,k}$ denote the $(i,j)\nth$ element of the $k\nth$ frontal slice. Denote by $\X_k$ the matrix comprising the $k\nth$ frontal slice. 
We use $\odot$ to denote the \define{Hadamard} (i.e., element-wise) product, $\tr(\cdot)$ for the trace operator and $\fronorm{\cdot}$ for the Frobenius norm. For a matrix $\mat V$ and function $f$, let $\grad_{\mat V} f$ denote the gradient of $f$ with respect to $\mat V$. 

Fix a set of $\nobj$ objects and a set of $\nrel$ relations.\footnote{Here, we use the term \emph{relation} loosely to include not only strict relations, for which relationships are either present or not, but also real-valued functions.} To simplify our analysis, we assume that all relations are symmetric, though one can obtain an analogous derivation for asymmetric relations with only slightly more work. We are given a partially observed tensor $\Y \in \Reals^{\nobj \times \nobj \times \nrel}$, in which each observed entry $\y_{i,j,k}$ is a (possibly noisy) measurement of a relationship and each unobserved entry is set to a null value.\footnote{For example, for binary-valued relations in $\TwoClass$, the null value is 0.}
We are additionally given a nonnegative weighting tensor $\W \in {\Reals^+}^{\nobj \times \nobj \times \nrel}$, where each entry $\w_{i,j,k} \in [0,1]$ corresponds to a user-defined confidence, or certainty, in the value of $\y_{i,j,k}$; if $\y_{i,j,k}$ is unobserved, then $\w_{i,j,k}$ is necessarily zero. The goal of multi-relational transduction in this tensor formulation is to infer the unobserved entries in $\Y$.

\section{Proposed Method}
\label{sec:method}
This section introduces our proposed method, which we refer to as \define{\Algorithm} (\Algo). 
We begin by describing our low-rank tensor representation of multi-relational data. We then define an optimization objective used to compute this representation and discuss how we solve the optimization.

\subsection{Representation as Tensor Decomposition}
\label{sec:model}

Our fundamental assumption is that each relationship is equal to a mapping $\trans_k$ applied to an element $\x_{i,j,k}$ in an underlying low-rank tensor $\X \in \Reals^{\nobj \times \nobj \times \nrel}$. Each $\trans_k$ depends on the nature of the relation, and may differ across relations. For example, for binary relations in $\TwoClass$, $\trans_k$ is the \emph{sign} function. We further assume that each $\X_k$ can be factored as a rank-$\nrank$ decomposition 
\begin{equation}
\label{eq:decomp}
\X_k = \A \R_k \A\T + \bias_k,
\end{equation}
where $\A \in \Reals^{\nobj \times \nrank}$, $\R_k \in \Reals^{\nrank \times \nrank}$ and $\bias_k \in \Reals$. (\autoref{fig:decomp} illustrates this decomposition.) Note that there is a single $\A$ matrix, but $\nrel$ instances of $\R_k$ and $\bias_k$. Also note that we place no constraints on $\A$ or $\R_k$; the columns of $\A$ need not be linearly independent, and $\R_k$ need not be positive-semidefinite. To infer the values of the missing (or uncertain) entries, we predict each $\y_{i,j,k}$ by computing $\x_{i,j,k} = \avec_i \R_k \avec_j\T + \bias_k$, where $\avec_i$ and $\avec_j$ are the $i\nth$ and $j\nth$ row vectors of $\A$, and then apply the appropriate mapping $\trans_k(\x_{i,j,k})$.

The entries of $\A$ can be interpreted as the \define{global latent factors} of the objects, where the $i\nth$ row $\avec_i$ corresponds to the latent factors of object $i$. Each $\R_k$ determines the interactions of $\A$ in the $k\nth$ relation. Thus, each predicted relationship comes from a linear combination of the objects' latent factors. Because the latent factors are global, information propagates between relations during the decomposition, thus enabling collective learning. The addition of $\bias_k$ accounts for distributional bias within each relation.

\subsection{Related Models}
\label{sec:related_work}

Our tensor model is comparable to \citeauthor{harshman:dedicom78}'s DEDICOM \citeyearpar{harshman:dedicom78}. \citet{bader:icdm07} applied the DEDICOM model to the task of temporal link prediction (in a single network), using the third mode as the time dimension. Recently, \citet{nickel:icml11} proposed a relaxed DEDICOM, referred to as RESCAL, to solve several canonical multi-relational learning tasks. Of the previous approaches, our underlying decomposition is most similar to RESCAL, and \autoref{eq:decomp} would be identical to the RESCAL decomposition if not for the bias term. Beyond the decomposition, the key distinction is that RESCAL directly decomposes the input tensor, rather than modeling the mapping from $\X$ to $\Y$. RESCAL also ignores the potential sparsity and uncertainty in the observations, whereas we explicitly model this. We demonstrate in \autoref{sec:experiments} that our formulation produces more accurate predictions even when observed (training) data is limited. 

Other tensor factorization models have been proposed for multi-relational data, though they typically use the CP decomposition \citep{dunlavy:sandia06,dunlavy:tkdd11,gao:cidm11,xiong:sdm10}. In the CP decomposition, each entry is the inner product of three vectors; this would be similar to our decomposition if each $\R_k$ slice were constrained to be diagonal. 
The richer interactions of the relaxed DEDICOM and the global latent representation of the objects often make it better suited for multi-relational learning, as was corroborated empirically by \citet{nickel:icml11}.

\subsection{Objective}
\label{sec:objective}
To compute the decomposition in \autoref{eq:decomp}, we minimize the following regularized objective:
\begin{align}
\label{eq:objfunc}
&\F(\A,\R,\biasvec) \defeq \frac{\reg}{2}\fronorm{\A}^2 \nonumber\\
	&~~~~ + \sum_{k=1}^{\nrel} \frac{\reg}{2}\fronorm{\R_k}^2
	+ \tr\( \W_k (\loss_k(\Y_k,\X_k))\T \),
\end{align}
where $\reg \geq 0$ is a regularization parameter, $\X_k$ is computed by \autoref{eq:decomp}, and $\loss_k$ is a loss function that is applied element-wise to the $k\nth$ slice. (For brevity, we use $\F$ to denote $\F(\A,\R,\biasvec)$.) This ability to combine multiple loss functions is central to our approach, as the appropriate penalty depends on the mapping for each $\X_k$ to $\Y_k$. Though most matrix and tensor decompositions focus on minimizing the quadratic loss (defined below), this criterion may not be optimal for certain prediction tasks (such as binary prediction). By explicitly making the loss function for each slice task-specific, our framework offers more flexibility than related techniques. The only requirement (due to our optimization method) is that the loss function is smooth.

It is important to note our use of L2 regularization. 
Regularization effectively controls the complexity of the model and thereby reduces the possibility of overfitting. This follows the traditional wisdom that ``simpler" models will generalize better to unseen data---in this case, the unobserved tensor entries. The rank of the decomposition can also be seen as a complexity parameter, since higher ranks will better fit the observed data. However, after a certain point, increasing the rank has a diminishing effect, since the regularizer seeks to minimize the Frobenius norm of the decomposition. We explore the effect of the rank parameter empirically in \autoref{sec:rank_exp}.

To minimize \autoref{eq:objfunc}, we require the gradients of $\F$ w.r.t.\ $\A$, $\R_k$ and $\bias_k$. Leveraging the symmetry of $\R_k$, we derive\footnote{Due to space restrictions, we state the gradients without their derivation.} these as
\begin{align}
\label{eq:gradA}
&\!\! \grad_{\A} \F
	= \reg \A + \! \sum_{k=1}^{\nrel} 2(\W_k \odot \grad_{\X_k}\loss_k(\Y_k,\X_k)) \A \R_k\T, \\
\label{eq:gradR}
&\!\! \grad_{\R_k} \F
	= \reg \R_k + \A\T \(\W_k \odot \grad_{\X_k}\loss_k(\Y_k,\X_k) \) \A, \\
\label{eq:gradb}
&\!\! \grad_{\bias_k} \F
	= \tr\( \W_k (\grad_{\X_k}\loss_k(\Y_k,\X_k))\T \),
\end{align}
where $\odot$ denotes the \define{Hadamard} (i.e., element-wise) product, and $\grad_{\X_k}\loss_k(\Y_k,\X_k)$ is the gradient of $\loss_k$ w.r.t.\ $\X_k$. Though this accommodates any differentiable loss function, we now present three that are applicable to many relational problems, and derive their corresponding loss gradients.

\paragraph{Quadratic Loss:}
The most common loss function used in matrix and tensor factorization is the \define{quadratic loss}, which we denote by $\qloss(\y,\x) \defeq \frac{1}{2} (\y-\x)^2$. Minimizing the quadratic loss corresponds to the setting in which each relationship is directly approximated by a linear combination of latent factors; i.e., $\trans_k$ is the identity and $\Y_k \approx \X_k$. For this loss function, the loss gradient is simply $\grad_{\X_k}\qloss_k(\Y_k,\X_k) = (\X_k - \Y_k)$.

\paragraph{Smooth Hinge Loss:}
While the quadratic loss may be appropriate for learning real-valued functions, it is sometimes ill-suited for learning binary relations, which are essentially binary classifications. For binary classification, the goal is to complete a partially observed slice $\Y_k \in \TwoClass^{\nobj \times \nobj}$. Recall that the mapping $\trans_k$ is the sign function, and so $\y_{i,j,k} \approx \sgn(\x_{i,j,k})$. Approximating $\TwoClass$ with a quadratic penalty may yield a ``small-margin" solution, since high-confidence predictions will push low-confidence predictions closer to the decision boundary. To get a ``large-margin" solution, we use the \define{smooth hinge loss} \citep{rennie:icml05}, $\hloss(\y,\x) \defeq \hinge( \y \x )$,
where
\begin{equation*}
\hinge(z) \defeq
	\begin{cases}
	1/2 - z & \text{if} ~ z \leq 0, \\
	(1 - z)^2 / 2 & \text{if} ~ 0 < z < 1, \\
	0 & \text{if} ~ z \geq 1.
	\end{cases}
\end{equation*}
Unlike the standard hinge loss, the smooth hinge is differentiable everywhere. To obtain closed-form gradients, we define tensors $\Pmat, \Qmat \in \Reals^{\nobj \times \nobj \times \nrel}$, where
\begin{equation*}
\p_{i,j,k} \defeq 
	\begin{cases}
	1 & \text{if} ~ 0 < \y_{i,j,k} \x_{i,j,k} < 1, \\
	0 & \text{otherwise},
	\end{cases}
\end{equation*}
and
\begin{equation*}
\q_{i,j,k} \defeq 
	\begin{cases}
	1 & \text{if} ~ \y_{i,j,k} \x_{i,j,k} < 1, \\
	0 & \text{otherwise}.
	\end{cases}
\end{equation*}
We can therefore express the smooth hinge as
\begin{equation*}
\hinge( \y_{i,j,k} \x_{i,j,k} ) = (\p_{i,j,k}\x_{i,j,k}^2 - 2\q_{i,j,k}\y_{i,j,k}\x_{i,j,k} + \q_{i,j,k}) / 2,
\end{equation*}
which we can differentiate w.r.t. $\X_k$ to obtain $\grad_{\X_k}\hloss_k(\Y_k,\X_k) = (\Pmat_k \odot \X_k - \Qmat_k \odot \Y_k)$.

\paragraph{Logistic Loss:}
For binary relations, we can also use the \define{logistic loss} \citep{rennie:ijcai05}, defined as $\lloss(\y,\x) \defeq \log(1+e^{-\y\x}))$. From a statistical perspective, this corresponds to the negative conditional log-likelihood of a logistic model. Note that this loss function also maximizes the binary prediction margin $\y\x$. The gradient of $\lloss$ is easily derived as 
$\grad_{\X_k}\lloss_k(\Y_k,\X_k) = -\Y_k\odot\Zmat_k$, where $\z_{i,j,k} \defeq (1+e^{\y\x})^{-1}$.

\subsection{Weighting and Efficiency}
\label{sec:weighting}

The weighting tensor $\W$ is a particularly important component of our framework. Without $\W$, the objective function would place equal importance on fitting both observed and unobserved values. If the observed tensor is very sparse (as it often is in real training data), this will result in fitting a large number of ``phantom zeros". The weighting tensor prevents this from happening by emphasizing only the observed (or certain) entries. We can thus train on a small number of observations without fitting the unobserved data. This approach is similar to \citeauthor{acar:sdm10}'s \citeyearpar{acar:sdm10}, though their analysis is limited to the minimizing the quadratic loss for a CP decomposition.

Weighting the objective by $\W$ also leads to an improvement in efficiency.
When $\W$ is sparse, the objective and gradient calculations are fairly lightweight, because any expression involving $\W$ can be computed using sparse arithmetic. 
For instance, $\X_k$ only appears in a Hadamard product with $\W_k$, so \autoref{eq:decomp} can be implemented as a sparse outer product, where we only compute $\x_{i,j,k}$ for any nonzero $\w_{i,j,k}$.
In Equations \ref{eq:objfunc}--\ref{eq:gradb}, the only expressions that do not involve $\W$ are the regularization terms. 
Thus, when $\W$ has only $\nnz$ nonzero elements, the computational costs of these equations are $\bigO(\nrel\nnz\nrank + \nrel\nobj\nrank^2)$. In contrast, methods that ignore the sparsity of the observed tensor take $\bigO(\nrel\nobj^2\nrank + \nrel\nobj\nrank^2)$ time. Assuming that $\nobj^2$ is the dominant term and that $\nnz$ grows much slower than $\nobj^2$ (e.g., in natural networks, $\nnz$ is often $\bigO(\nobj)$), the sparse computation can be an order of magnitude faster.

Additionally, since $\W$ can be real-valued (not just $\Binary$), we can adjust the entries to reduce the mistake penalty of certain examples. For instance, suppose an incorrect negative prediction is deemed more critical than an incorrect positive (as is often the case in medical diagnoses and certain link prediction tasks). One could multiplicatively increase the values of all $\{ \w_{i,j,k} : \y_{i,j,k}=1 \}$ or, alternatively, decrease the values of all $\{ \w_{i,j,k} : \y_{i,j,k}=-1 \}$. This would effectively penalize false negatives more severely than false positives, encouraging the optimization to satisfy positive examples.

\subsection{Optimization}
\label{sec:optimization}

To minimize the objective in \autoref{eq:objfunc}, we use \define{limited-memory Broyden-Fletcher-Goldfarb-Shanno} (L-BFGS) optimization. Since quasi-Newton methods, such as L-BFGS, avoid computing the Hessian, they are efficient for optimization problems involving many variables.
All this requires is the objective function and the gradients in Equations \ref{eq:gradA}--\ref{eq:gradb}.
Since our optimization problem is non-convex, we are not guaranteed that L-BFGS will find the global minimum; in practice, however, the algorithm typically finds useful, though possibly local, minima.


To mitigate the possibility of finding local minima, we initialize the parameters using the eigendecomposition of each input slice, which is close to the desired factorization. This technique is similar to the initialization used by \citet{bader:icdm07} and \citet{nickel:icml11}, which have similar decompositions. For $k = 1,\dots,\nrel$, let $\eigvals_k \defeq (\eigval_{1,k},\dots,\eigval_{\nrank,k})$ denote the $\nrank$ largest eigenvalues of $\Y_k$, and let $\eigvecs_k \defeq (\eigvec_{1,k},\dots,\eigvec_{\nrank,k})$ denote their corresponding eigenvectors. We initialize $\R_k$ as a diagonal matrix with $\eigvals_k$ along the diagonal, and $\A$ as the average of $\eigvecs_1,\dots,\eigvecs_{\nrel}$. In practice, we find that this initialization converges faster, and often to a better solution, than random initialization.


Note that when the objective function uses only quadratic loss, one can compute the parameter updates using the \define{alternating simultaneous approximation, least squares and Newton} (ASALSAN) algorithm \citep{bader:icdm07}, which produces an approximate solution and has been shown to converge quickly. Since our objective may contain a heterogeneous mixture of loss functions---not all necessarily quadratic---we do not use ASALSAN. Morever, we cannot use traditional convex programming techniques like \emph{semidefinite programming} (SDP) because our objective is non-convex. 

\section{Experiments}
\label{sec:experiments}

In this section, we compare variants of \Algo\ with RESCAL \citep{nickel:icml11}, MMMF \citep{rennie:icml05} and \define{Bayesian probabilistic tensor factorization} (BPTF) \citep{xiong:sdm10} in several experiments, using both real and synthetic data. The real data sources are kinship data from the Australian Alyawarra tribe, and two social interaction datasets from the MIT Media Lab. The comparisons highlight the critical advantages of \Algo: namely, the ability to learn from limited training data, handle a mixture of learning objectives, transfer information across relations for collective learning, and exploit sparsity for improved efficiency.

To test the effect of the rank parameter, we run an experiment varying only the rank of the decomposition over a range of values. The results support our hypothesis that L2 regularization reduces the impact of the rank, effectively controlling the model complexity.

Finally, we perform a synthetic experiment to compare the running time of \Algo\ to that of the above competing methods, demonstrating the significant scalability gains provided by exploiting sparsity.

To conserve space, certain figures and tables are provided in the supplementary material (\autoref{sec:appendix}).

\subsection{Compared Methods}
\label{sec:compared_methods}

To evaluate the performance of various loss functions, we compare several variants of \Algo. The variant named \Algo-Q uses the quadratic loss for all relations, regardless of their type. \Algo-H and \Algo-L use the quadratic loss for real-valued slices and the smooth hinge or logistic loss, respectively, for binary slices.

The RESCAL model approximates each slice of the input tensor as $\Y_k \approx \A \R_k \A\T$. In \citep{nickel:icml11}, binary relationships are represented by $\Binary$. Unfortunately, since RESCAL does not account for missing data, unobserved relationships are simply treated as negative examples. In order to distinguish between (un)observed relationships and negative examples, we use $\TwoClass$ for observed data and zeros elsewhere. In our experiments, we find that this modification improves RESCAL's performance over the original method. Since RESCAL uses the quadratic loss uniformly, it uses ASALSAN to compute the decomposition, with L2 regularization on $\A$ and $\R_k$.

MMMF is a tool for matrix reconstruction and, as such, is not designed for multi-relational data. That said, we can use it to reconstruct each slice of the tensor individually. Like \Algo, MMMF approximates a binary input $\Y$ using the sign of a rank-$\nrank$ matrix decomposition, $\Y \approx \sgn(\mat U \mat V\T)$, where $\mat U, \mat V \in \Reals^{\nobj \times \nrank}$. The ``fast" variant of the algorithm \citep{rennie:icml05} adds a bias term and uses the smooth hinge loss. The optimization objective is very similar to ours, but with different gradients, due to the decomposition. Our implementation of fast MMMF differs from that of \citet{rennie:icml05} only in the way we solve the optimization (using L-BFGS, rather than conjugate gradient descent) and the fact that the input is assumed to be symmetric.

Because the synthetic data generator (described in \autoref{sec:synth_exp}) matches our decomposition and is slightly different than that of traditional MMMF, it is somewhat unfair to compare traditional MMMF to \Algo. We therefore run a variant of \Algo\ that decomposes each slice separately instead of jointly, using a separate $\A_k$. This is meant to equalize the discrepancy in the decomposition, while isolating the deficiencies of non-collective learning. We refer to this model as MMMF+.

BPTF is a fully Bayesian interpretation of the CP tensor factorization, originally designed for temporal prediction.\footnote{\citet{sutskever:nips09} propose another fully Bayesian algorithm, \define{Bayesian tensor factorization} (BTF), whose decomposition is very similar to ours, though their framework only supports the quadratic loss. We were unable to compare \Algo\ to this method at the time of submission.} We compare it to \Algo\ to investigate the benefits and drawbacks of the Bayesian approach. BPTF assumes that all latent factors are sampled from a Gaussian distribution. The only user-defined properties are the rank of the decomposition and the hyper-parameters. The parameters and latent factors are estimated using Gibbs sampling. One benefit of the Bayesian approach is that it avoids the model selection problem, which in our case is the choice of regularization parameters.\footnote{As the authors claim, the effect of tuning the hyper-parameter priors is minimal.} This can have a pronounced effect when training data is limited, which makes proper regularization critical. However, Gibbs sampling is computationally expensive, since it requires many iterations of sampling to converge to an accurate estimate. We analyze this trade-off between accuracy and efficiency in \autoref{sec:timing}. Additionally, BPTF only supports the quadratic loss, since it has a natural probabilistic interpretation as the Gaussian likelihood and makes the model conjugate, making Gibbs sampling easier. No such interpretation exists for the (smooth) hinge loss, and the logistic loss has no conjugate prior.

We implement all of the above methods in MATLAB, using a third-party implementation of L-BFGS\footnote{\url{www.di.ens.fr/~mschmidt/Software/minFunc.html}}, and the authors' implementation of BPTF\footnote{\url{www.cs.cmu.edu/~lxiong/bptf/bptf.html}}.

\subsection{Synthetic Data Experiments}
\label{sec:synth_exp}

\begin{figure*}[tb]
\begin{center}
\subfigure[\SynthB]{\includegraphics[scale=0.265]{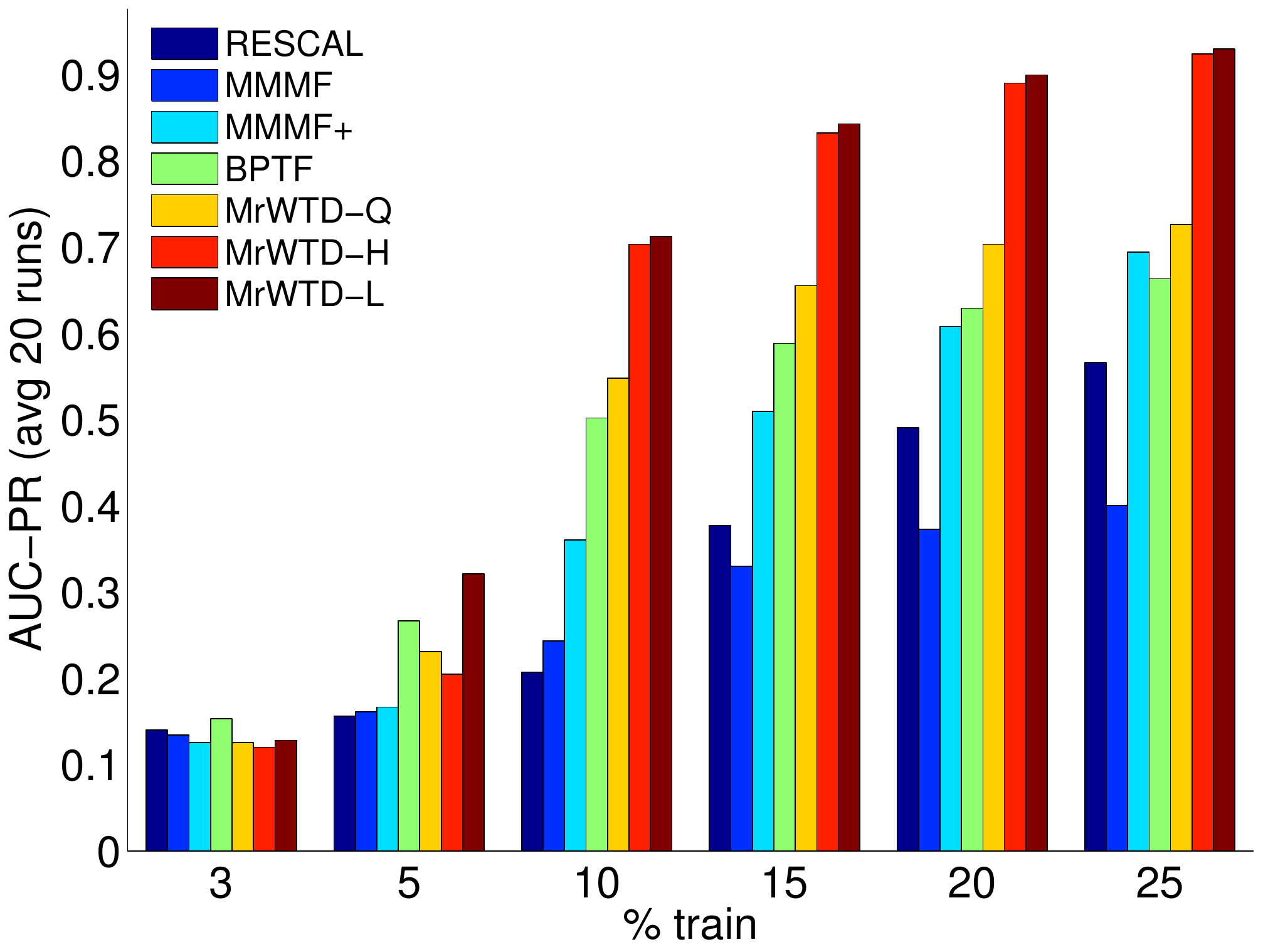}} ~
\subfigure[\SynthM\ (binary slice)]{\includegraphics[scale=0.265]{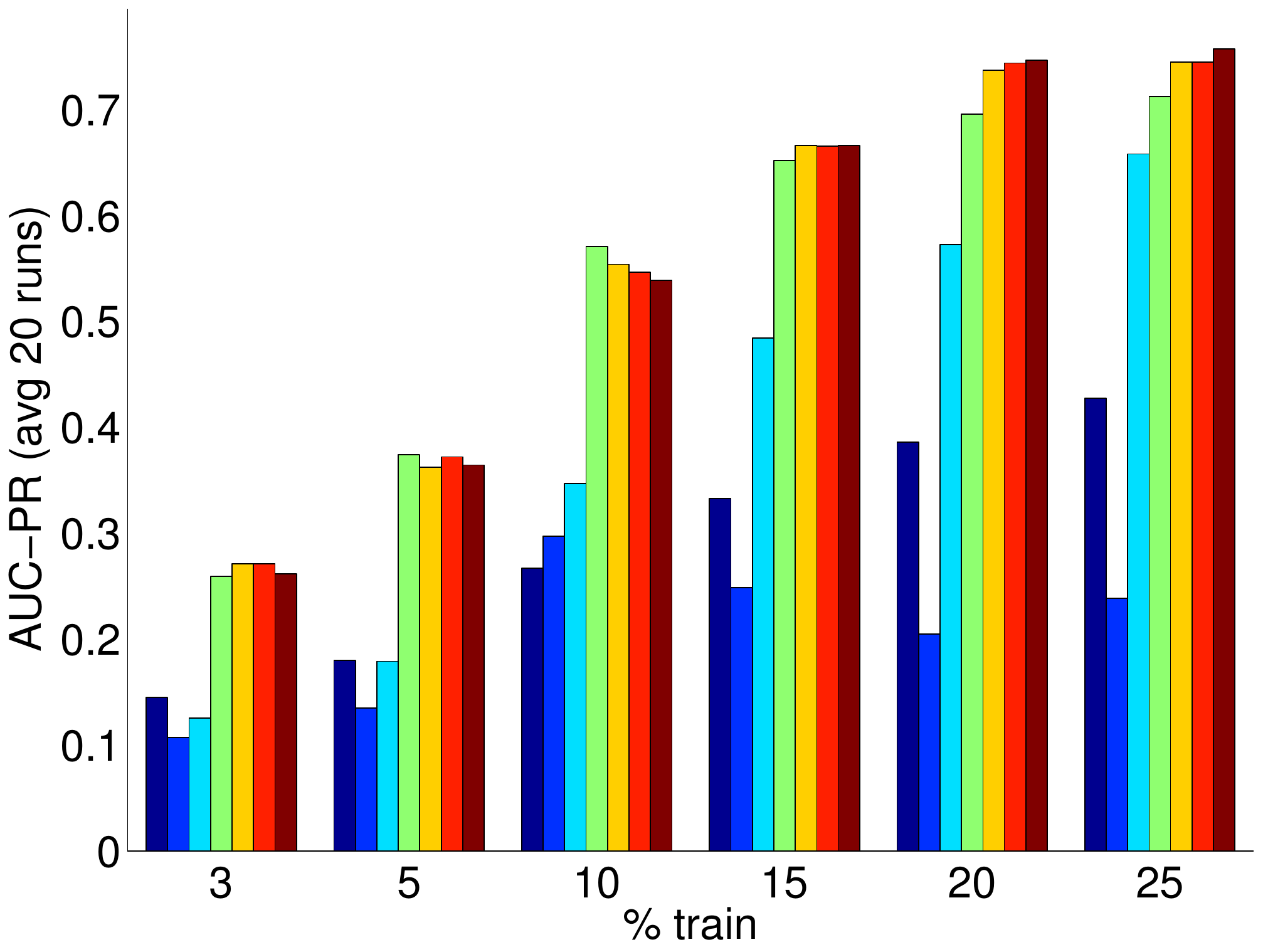}} ~
\subfigure[\SynthM\ (real slice)]{\includegraphics[scale=0.265]{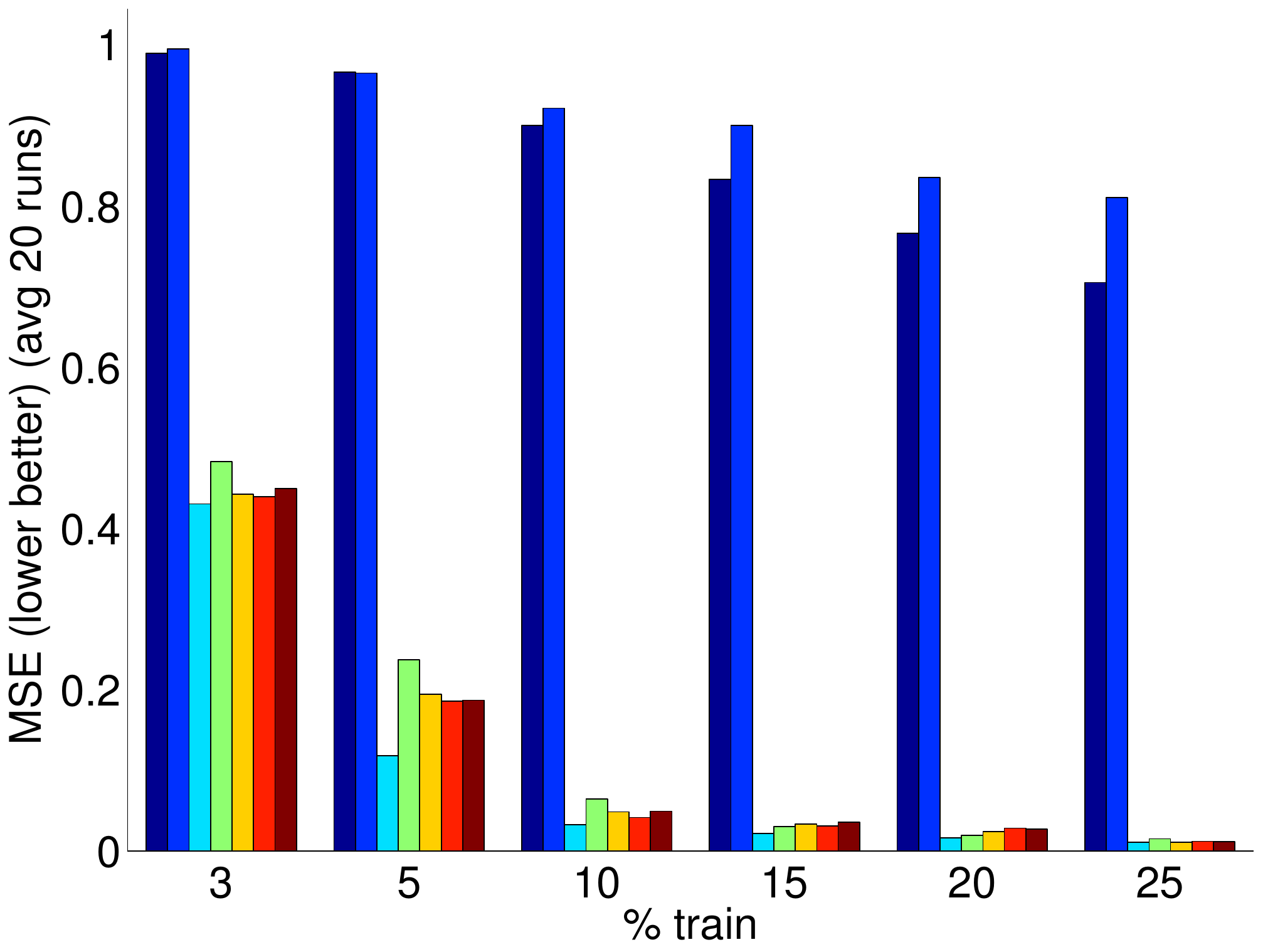}}
\caption{Results of the synthetic data experiments (discussed in \autoref{sec:synth_exp}). The horizontal axis indicates the size of the training data (in percentage of the tensor); in (a) and (b), the vertical axis indicates the area under the precision-recall curve; in (c), the vertical axis shows the mean squared error (MSE). All scores are averaged over 20 runs. The top to bottom arrangement of the legend corresponds to a left to right arrangement in each group.
See \autoref{tab:synth_results} in the appendix for the full results, including standard deviations.}
\label{fig:synth_results}
\end{center}
\end{figure*}

To generate the synthetic data, we start by computing a low-rank tensor $\hat\X \in \Reals^{\nobj \times \nobj \times \nrel}$ as $\hat\X_k \gets \hat\A \hat\R_k \hat\A\T + \Noise_k$, for $k=1,\dots,\nrel$, where $\hat\A \in \Reals^{\nobj \times \nrank}$ and $\hat\R_k \in \Reals^{\nrank \times \nrank}$ are sampled from a normal distribution, and $\Noise_k \in \Reals^{\nobj \times \nobj}$ is low-level, normally-distributed noise. For the first experiment, we construct $\nrel=3$ binary relations (i.e., slices), over $\nobj=500$ objects, using rank $\nrank=10$. We refer to this dataset as \SynthB. To generate a binary tensor $\Y \in \TwoClass^{\nobj \times \nobj \times \nrel}$, we round the values of $\hat\X$ using the $90\nth$ percentile of its values as a threshold. This produces a heavy skew towards the negative class, as is typical in real multi-relational data. For the second experiment, we construct one binary relation and one real-valued relation, again over 500 objects, with rank 10. We normalize the real-valued relation such that the standard deviation is 1.0, giving it roughly the same scale as the binary slices. We refer to this dataset as \SynthM.

We evaluate over training sizes $t \in [3,25]$ percent, averaging the results over 20 runs per size. In each run, we sample a random $t \cdot {\nobj \choose 2}$ pairs (and their symmetric counterparts) from each slice to use as the training set, and let the remaining pairs comprise the test set. We then hold out a random 25\% from the training set as a validation set for a regularization parameter search, where we search over the range $[10^{-3},10^3]$ in logarithmic increments. For \SynthB, we select the optimal parameter $\reg^*$ that maximizes the area under the precision-recall curve (AUPRC), averaged over all slices; for \SynthM, we maximize the harmonic mean of the AUPRC of the first slice and one minus the mean-squared error (MSE) of the second. We then retrain on the full training set using $\reg^*$ and evaluate on the test set. For BPTF, we run Gibbs sampling for 200 iterations.

The results of the synthetic data experiments are given in \autoref{fig:synth_results}, reported as average AUPRC and MSE over 20 runs. 
On \SynthB, \Algo-L achieves a statistically significant\footnote{We measure statistical significance in all experiments using a 2-sample t-test with rejection threshold 0.05.} lift over the competing methods for training sizes 5\% and up, and all three variants showing significant lift for 10\% and above. We attribute these results to two primary advantages: the weighted objective function, with its mixture of task-specific loss functions, and the global latent factors. As discussed in \autoref{sec:optimization}, the weighted objective is necessary for exploiting small amounts of observed (i.e., training) data, without fitting the unobserved entries. Since RESCAL treats all entries as observed, it tends to fit the unobserved entries in sparsely populated tensors. Furthermore, though MMMF and MMMF+ use the same large-margin technique as \Algo-H and \Algo-L, they do not perform collective learning, since the latent factors are specific to each slice. In \Algo, information from one slice is propagated to the others via the global latent factors. Note that BPTF and \Algo-Q perform significantly worse the large-margin loss variants of \Algo\ for sizes 10\% and above, illustrating that the quadratic loss is not always appropriate for binary data.
On \SynthM, \Algo's improvement over RESCAL and MMMF, for both slices, is statistically significantly for all training sizes. 
MMMF+ is competitive with \Algo\ on the real-valued slice, with significant lift for training sizes $3,5$\%; yet its performance deteriorates on the binary slice, for all training sizes, since it is not able to transfer information between slices. BPTF is also competitive with \Algo\ on the real-valued slice, and the binary slice for smaller training sizes, but falls slightly behind on the higher sizes.

\subsection{Real Data Experiments}
\label{sec:real_exp}

We evaluate on several real multi-relational datasets.
%
The first dataset consists of kinship data from the Australian Alyawarra tribe, as recorded by \citet{denham:fm05}. This data has previously been used by \citet{kemp:aaai06} for multi-relational link prediction. The data contains $\nobj=104$ tribe members and $\nrel=23$ types of kinship (binary) relations.\footnote{The original data contains 26 relations, but relations 24--26 are extremely sparse, exhibiting fewer than 6 instances, so we omit them.} In total, the dataset includes 125,580 related pairs. This yields a tensor $\Y \in \TwoClass^{104\times104\times23}$.

The remaining datasets come from MIT's Human Dynamics Laboratory. Both consist of human interaction data from students, faculty, and staff working on the MIT campus, recorded by a mobile phone application. From the first dataset, named Reality Mining \citep{eagle:nas09}, we use the survey-annotated network, consisting of $\nrel=3$ types of binary relationships annotated by the subjects: friendship, in-lab interaction and out-of-lab interaction. These relationships are measured between $\nobj=94$ participants, providing a total of 13,395 related pairs. The resulting tensor is $\Y \in \TwoClass^{94\times94\times3}$. From the second dataset, named Social Evolution \citep{dong:mum11}, we use the survey-annotated network, as well as several interaction relations derived from sensor data, resulting in $\nrel = 8$ binary relations with 16,101 related pairs. The five surveyed relations are: close friendship, biweekly social interaction, political discussion, two types of social media interaction. The three derived relations are computed from: voice calls, SMS messaging and proximity. We binarize this data by a simple indicator of whether the given type of interaction occurred. In this case, the number of users is $\nobj=84$, results in a tensor $\Y \in \TwoClass^{84\times84\times8}$.

\begin{figure*}[tb]
\begin{center}
\subfigure[Kinship]{\includegraphics[scale=0.265]{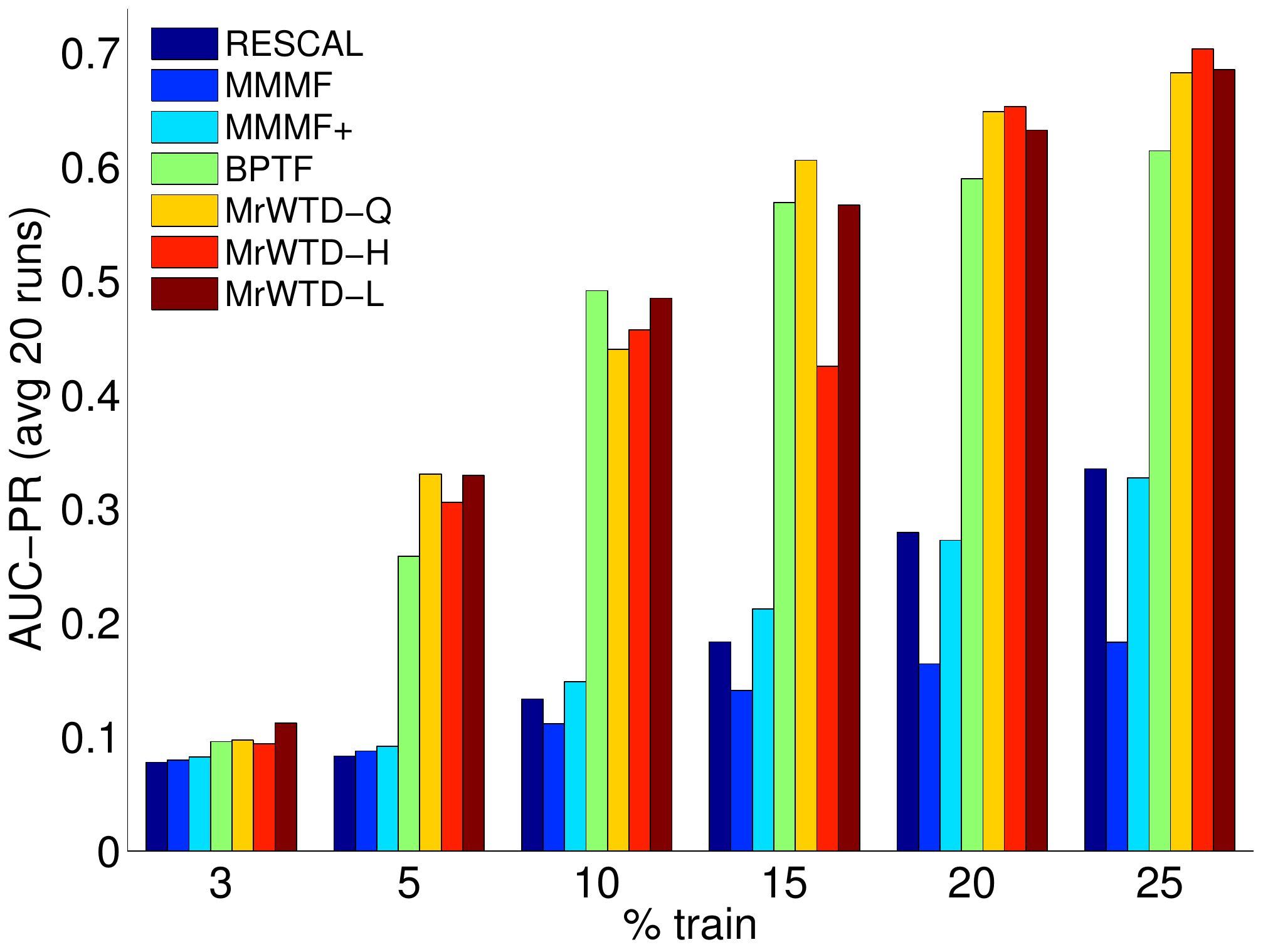}} ~
\subfigure[Reality Mining]{\includegraphics[scale=0.265]{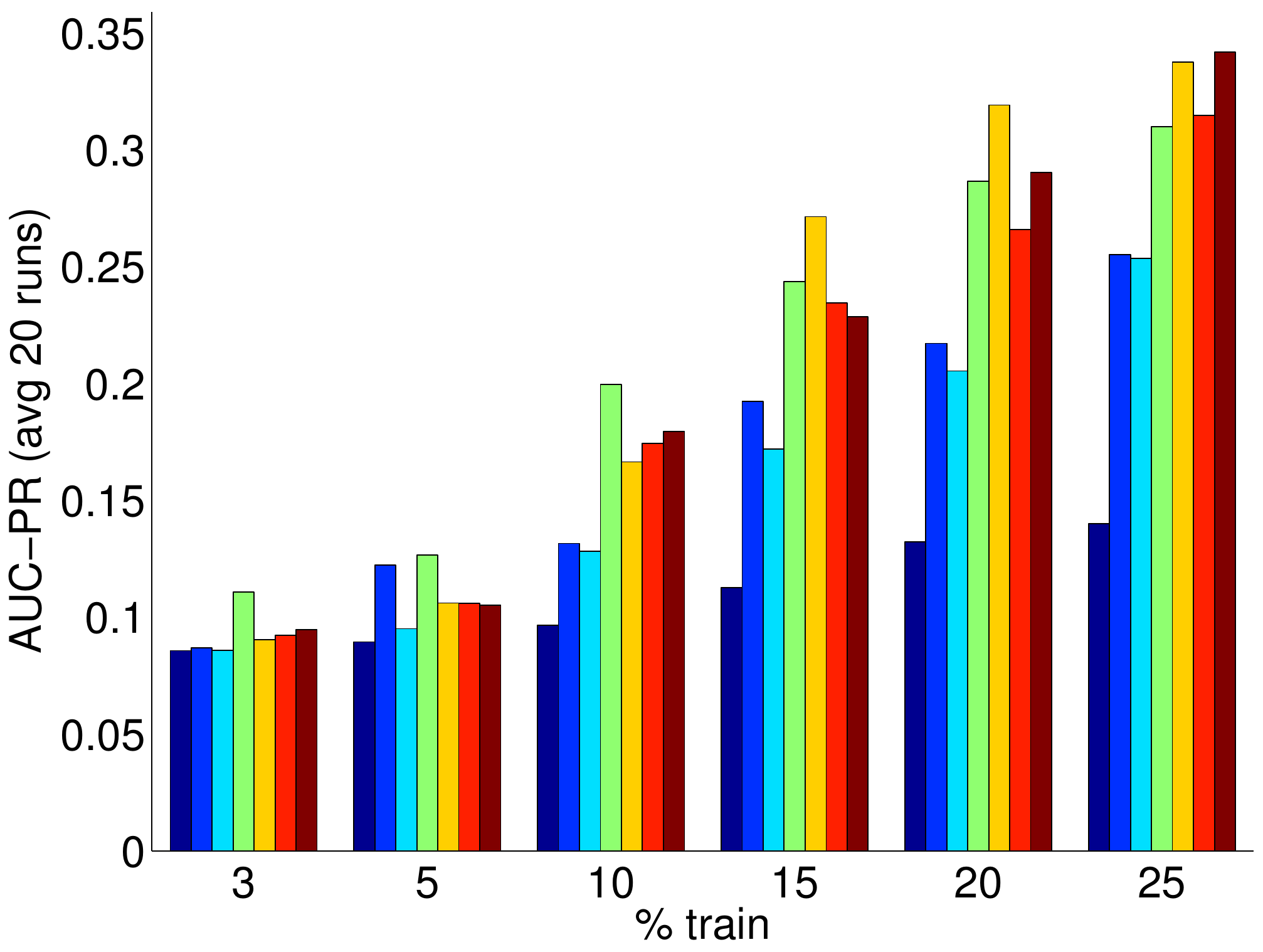}} ~
\subfigure[Social Evolution]{\includegraphics[scale=0.265]{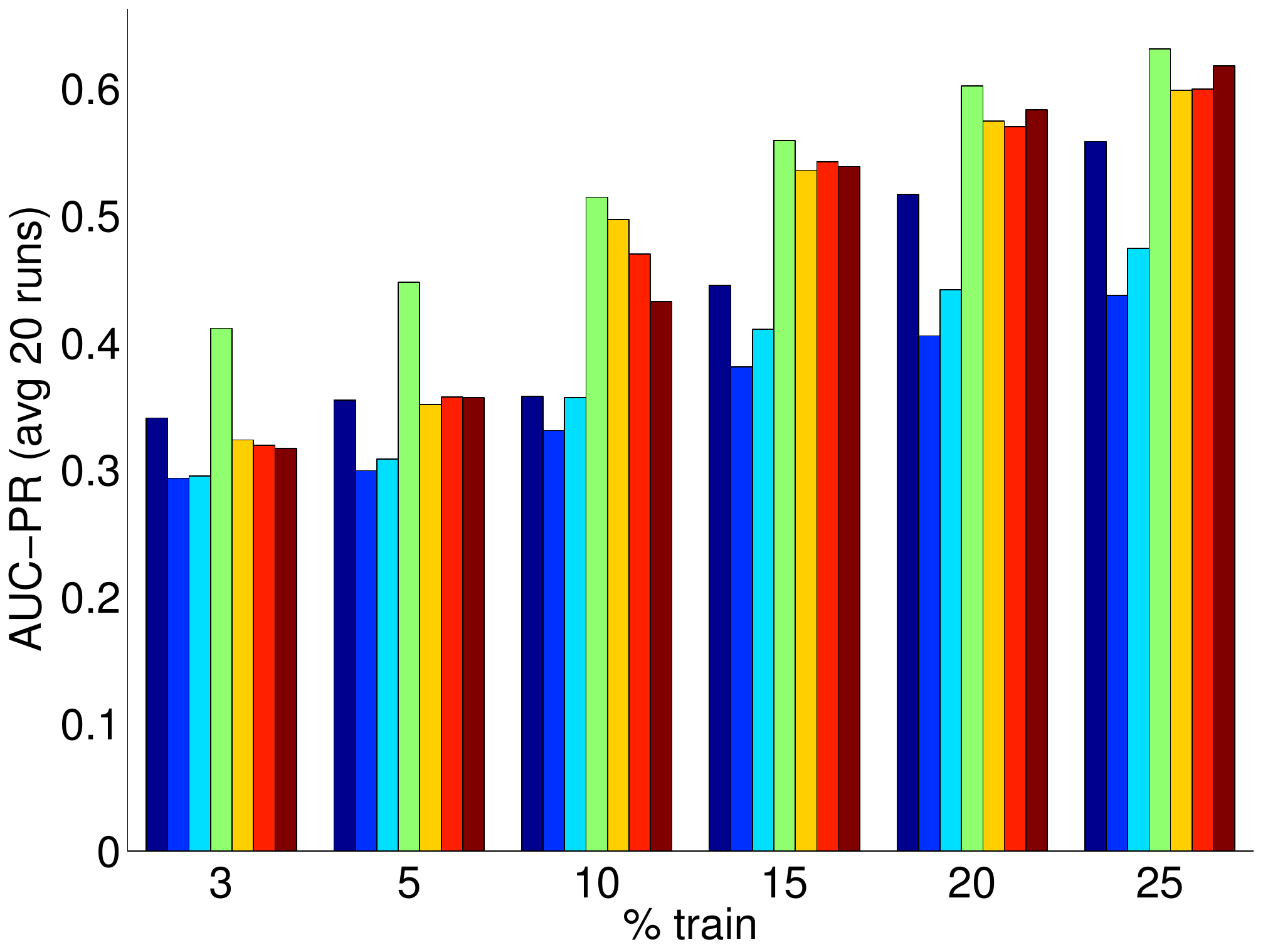}}
\caption{Results of the real data experiments (discussed in \autoref{sec:real_exp}). The horizontal axis indicates the size of the training data (in percentage of the tensor); the vertical axis indicates the area under the precision-recall curve, averaged over 20 runs. The top to bottom arrangement of the legend corresponds to a left to right arrangement in each group. See \autoref{tab:real_results} in the appendix for the full results, including standard deviations.}
\label{fig:real_results}
\end{center}
\end{figure*}

For these experiments, we use the same methodology as the synthetic experiments, with rank $\nrank = 20$. The results are also given in \autoref{fig:real_results}. 
The three variants of \Algo\ and BPTF achieve significant lift over RESCAL, MMMF and MMMF+ in nearly all experiments. \Algo\ has a statistically significant advantage over the other methods for most training ratios on the Kinship data, while BPTF has an advantage on the Social Evolution data.
Yet, as we show in the following section, \Algo's estimation takes a small fraction of BPTF's running time. We therefore achieve results that are comparable to Bayesian methods in far less time.
We refer the reader to \autoref{tab:timing} in the appendix for the complete set of results.

\subsection{Rank Experiment}
\label{sec:rank_exp}

To measure the effect of the rank parameter on the performance of each algorithm, we rerun the Social Evolution experiment, varying $\nrank = \{5,10,20,40\}$ and keeping the training ratio is fixed at 25\%. The results of this experiment are displayed in \autoref{fig:rank_results}, in the appendix. There is a small increase in AUC from $\nrank=5$ to $\nrank=10$, which is expected, since 5 is relatively low. However, we find that the effect of the rank is minimal for $\nrank\geq10$; the standard deviation across all runs in this range is $< 0.02$ for each algorithm. This supports our hypothesis that, beyond a certain threshold, the regularizer is the primary controller of model complexity.

\subsection{Timing Experiment}
\label{sec:timing}

Finally, we measure the running time of each of the above tensor methods to better understand their scalability in scenarios where training data is limited. We create a sequence of synthetic datasets (using the technique in \autoref{sec:synth_exp}), each with $\nrel = 3$ binary slices, for sizes $\nobj = \{500,1000,2000,4000,8000\}$. For training, we use a random 10\% of the tensor. We compare the smooth hinge loss variant of \Algo, RESCAL and BPTF, using predefined regularization and hyper-parameters. We run these experiments on a machine with two 6-core Intel\textsuperscript{\textregistered} Xeon\textsuperscript{\textregistered} X5650 processors, running at 2.66 GHz, and 48 GB of RAM.

\begin{figure}[ht]
\begin{center}
\includegraphics[scale=0.4]{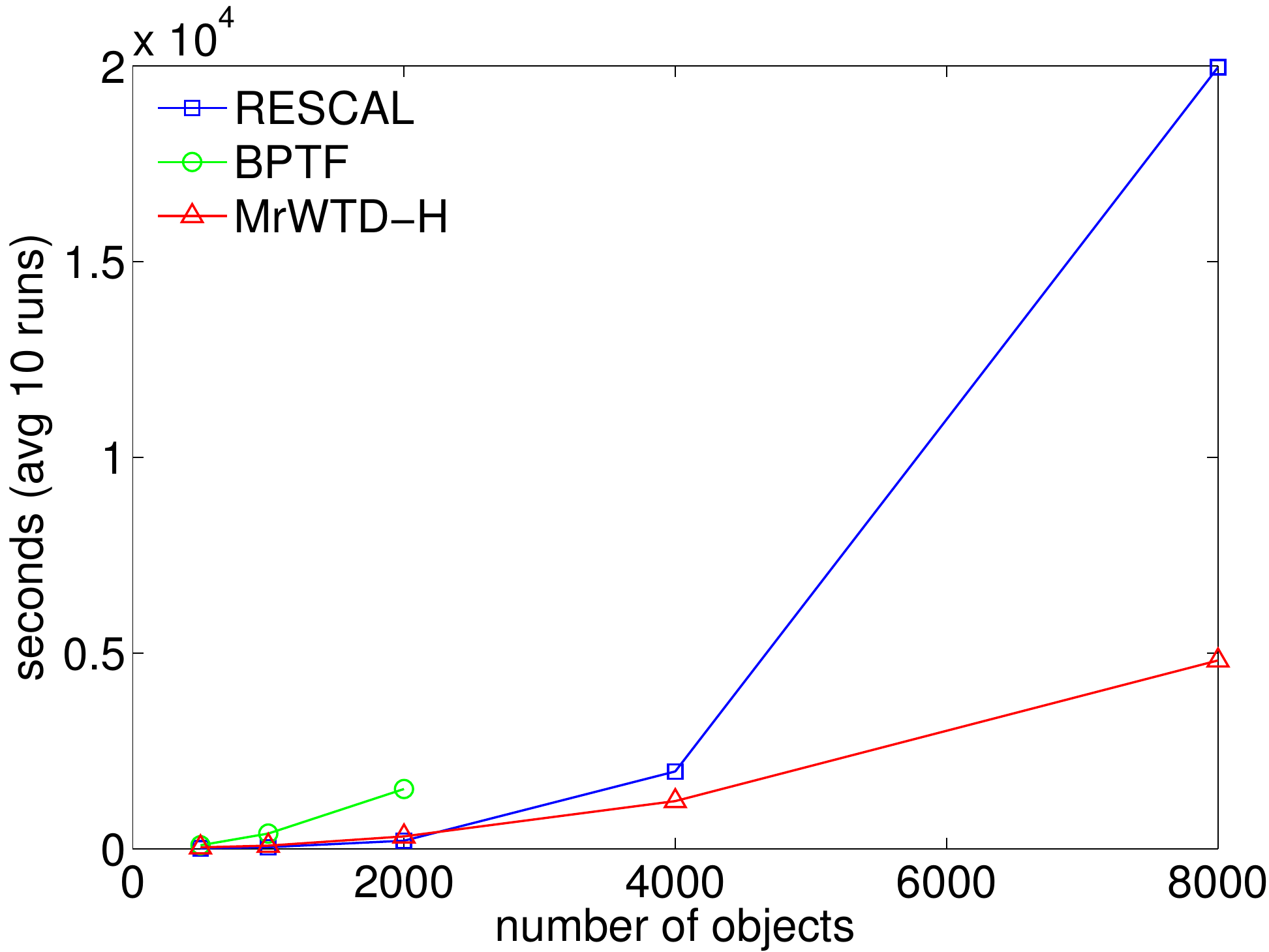}
\caption{Results of the timing experiment (discussed in \autoref{sec:timing}). The  horizontal axis indicates the size of a synthetic dataset, measured by the number of objects $\nobj$; the vertical axis indicates the running time (in seconds), averaged over 10 runs. For each dataset, we use 10\% for training. Note that BPTF could not run on sizes $\{4000,8000\}$ due to runtime exceptions.}
\label{fig:timing}
\end{center}
\end{figure}

The timing results, averaged over 10 runs per problem size, are shown in \autoref{fig:timing}. BPTF takes considerably more time than the others, due to its Gibbs sampling estimation. Note that we could not run BPTF on the two largest problem sizes, due to out-of-memory exceptions. This illustrates the tradeoff between accuracy and efficiency in using Bayesian methods; one can reduce running time by reducing the number of iterations, but this would also affect the accuracy of the estimation. Due to the efficient, closed-form updates of the ASALSAN algorithm, RESCAL is the fastest for small problem sizes. However, \Algo\ is significantly faster as the problem size grows. This is because RESCAL's objective function treats all tensor entries with equal importance, whereas \Algo's weighted objective only requires the predictions of the observed entries, thus allowing us to skip prediction on the test data during estimation.

\section{Conclusion}
\label{sec:conclusion}

In this paper, we present a modular framework for multi-relational learning via tensor decomposition. 
The decomposition we use provides an intuitive interpretation for the multi-relational domain, where objects have global latent representations and relationships are determined by a function of their linear combinations. 
We show that the global latent representations enable information to transfer between relation types during model estimation. 
Further, we demonstrate that our framework's weighted objective and support for multiple loss functions improves accuracy over similar models.
Finally, we show that our method exploits the sparsity of limited training data to achieve an order of magnitude speedup over unweighted methods.\looseness=-1

We plan to extend \Algo\ to be able to learn from large-scale data by adapting hashing methods from matrix factorization literature \citep{karatzoglou:aistats10}. We would also like to compare our method to \citeauthor{sutskever:nips09}'s BTF algorithm \citeyearpar{sutskever:nips09}, to further investigate the benefit of the Bayesian approach. We also intend to analyze the theoretical properties of our framework, such as generalization error, using existing learning theory literature
\citep{srebro:nips05b,cortes:icml08,elyaniv:jair09}.

\subsection*{Acknowledgements}
This work was partially supported by NSF CAREER grant 0746930 and NSF grant IIS1218488.

\bibliographystyle{plainnat}
\bibliography{mrwtd} 

\begin{thebibliography}{21}
\providecommand{\natexlab}[1]{#1}
\providecommand{\url}[1]{\texttt{#1}}
\expandafter\ifx\csname urlstyle\endcsname\relax
  \providecommand{\doi}[1]{doi: #1}\else
  \providecommand{\doi}{doi: \begingroup \urlstyle{rm}\Url}\fi

\bibitem[Acar et~al.(2010)Acar, Dunlavy, Kolda, and M{\o}rup]{acar:sdm10}
E.~Acar, D.~Dunlavy, T.~Kolda, and M.~M{\o}rup.
\newblock Scalable tensor factorizations with missing data.
\newblock In \emph{Proc. of the 2010 SIAM International Conf. on Data Mining
  (SDM)}, 2010.

\bibitem[Bader et~al.(2007)Bader, Harshman, and Kolda]{bader:icdm07}
B.~Bader, R.~Harshman, and T.~Kolda.
\newblock Temporal analysis of semantic graphs using {ASALSAN}.
\newblock In \emph{Proc. of the 7th IEEE International Conf. on Data Mining
  (ICDM)}, 2007.

\bibitem[Cortes et~al.(2008)Cortes, Mohri, Pechyony, and
  Rastogi]{cortes:icml08}
C.~Cortes, M.~Mohri, D.~Pechyony, and A.~Rastogi.
\newblock Stability of transductive regression algorithms.
\newblock In \emph{Proc. of the 25th International Conf. on Machine Learning
  (ICML)}, 2008.

\bibitem[Denham and White(2005)]{denham:fm05}
W.~Denham and D.~White.
\newblock Multiple measures of {A}lyawarra kinship.
\newblock \emph{Field Methods}, 17\penalty0 (1), 2005.

\bibitem[Dong et~al.(2011)Dong, Lepri, and Pentland]{dong:mum11}
W.~Dong, B.~Lepri, and A.~Pentland.
\newblock Modeling the co-evolution of behaviors and social relationships using
  mobile phone data.
\newblock In \emph{Proceedings of the 10th International Conference on Mobile
  and Ubiquitous Multimedia}, MUM '11, pages 134--143, New York, NY, USA, 2011.
  ACM.

\bibitem[Dunlavy et~al.(2006)Dunlavy, Kolda, and Kegelmeyer]{dunlavy:sandia06}
D.~Dunlavy, T.~Kolda, and W.~Kegelmeyer.
\newblock Multilinear algebra for analyzing data with multiple linkages.
\newblock Technical Report, 2006.

\bibitem[Dunlavy et~al.(2011)Dunlavy, Kolda, and Acar]{dunlavy:tkdd11}
D.~Dunlavy, T.~Kolda, and E.~Acar.
\newblock Temporal link prediction using matrix and tensor factorizations.
\newblock \emph{ACM Trans. on Knowledge Discovery from Data}, 5\penalty0 (2),
  2011.

\bibitem[Eagle et~al.(2009)Eagle, Pentland, and Lazer]{eagle:nas09}
N.~Eagle, A.~Pentland, and D.~Lazer.
\newblock Inferring friendship network structure by using mobile phone data.
\newblock \emph{Proc. of the National Academy of Sciences}, 106\penalty0 (36),
  2009.

\bibitem[El-Yaniv and Pechyony(2009)]{elyaniv:jair09}
R.~El-Yaniv and D.~Pechyony.
\newblock Transductive {R}ademacher complexity and its applications.
\newblock \emph{J. Artificial Intelligence Research (JAIR)}, 35, 2009.

\bibitem[Gao et~al.(2011)Gao, Denoyer, and Gallinari]{gao:cidm11}
S.~Gao, L.~Denoyer, and P.~Gallinari.
\newblock Link pattern prediction with tensor decomposition in multi-relational
  networks.
\newblock In \emph{IEEE Symposium on Comp. Intell. and Data Mining}, 2011.

\bibitem[Harshman(1978)]{harshman:dedicom78}
R.~Harshman.
\newblock Models for analysis of asymmetrical relationships.
\newblock In \emph{First Joint Meeting of the Psychometric Society and the
  Society for Mathematical Psychology}, 1978.

\bibitem[Karatzoglou et~al.(2010)Karatzoglou, Smola, and
  Weimer]{karatzoglou:aistats10}
A.~Karatzoglou, A.~Smola, and M.~Weimer.
\newblock Collaborative filtering on a budget.
\newblock In \emph{Proc. of the 13th International Conf. on Artificial
  Intelligence and Statistics}, 2010.

\bibitem[Kashima et~al.(2009)Kashima, Kato, Yamanishi, Sugiyama, and
  Tsuda]{kashima:sdm09}
H.~Kashima, T.~Kato, Y.~Yamanishi, M.~Sugiyama, and K.~Tsuda.
\newblock Link propagation: a fast semi-supervised learning algorithm for link
  prediction.
\newblock In \emph{SIAM International Conference on Data Mining (SDM)}, 2009.

\bibitem[Kemp et~al.(2006)Kemp, Tenenbaum, Griffiths, Yamada, and
  Ueda]{kemp:aaai06}
C.~Kemp, J.~Tenenbaum, T.~Griffiths, T.~Yamada, and N.~Ueda.
\newblock Learning systems of concepts with an infinite relational model.
\newblock In \emph{Proc. of the 21st National Conf. on Artificial
  Intelligence}, 2006.

\bibitem[Nickel et~al.(2011)Nickel, Tresp, and Kriegel]{nickel:icml11}
M.~Nickel, V.~Tresp, and H.~Kriegel.
\newblock A three-way model for collective learning on multi-relational data.
\newblock In \emph{Proc. of the 28th International Conf. on Machine Learning
  (ICML)}, 2011.

\bibitem[Rennie and Srebro(2005{\natexlab{a}})]{rennie:icml05}
J.~Rennie and N.~Srebro.
\newblock Fast maximum margin matrix factorization for collaborative
  prediction.
\newblock In \emph{In Proc. of the 22nd International Conf. on Machine Learning
  (ICML)}, 2005{\natexlab{a}}.

\bibitem[Rennie and Srebro(2005{\natexlab{b}})]{rennie:ijcai05}
J.~Rennie and N.~Srebro.
\newblock Loss functions for preference levels: regression with discrete
  ordered labels.
\newblock In \emph{IJCAI Multidisciplinary Workshop on Adv. in Preference
  Handling}, 2005{\natexlab{b}}.

\bibitem[{Srebro} et~al.(2005{\natexlab{a}}){Srebro}, {Alon}, and
  {Jaakkola}]{srebro:nips05b}
N.~{Srebro}, N.~{Alon}, and T.~{Jaakkola}.
\newblock Generalization error bounds for collaborative prediction with
  low-rank matrices.
\newblock In \emph{Advances in Neural Information Processing Systems 17}.
  2005{\natexlab{a}}.

\bibitem[{Srebro} et~al.(2005{\natexlab{b}}){Srebro}, {Rennie}, and
  {Jaakkola}]{srebro:nips05a}
N.~{Srebro}, J.~{Rennie}, and T.~{Jaakkola}.
\newblock Maximum-margin matrix factorization.
\newblock In \emph{Advances in Neural Information Processing Systems 17}.
  2005{\natexlab{b}}.

\bibitem[Sutskever et~al.(2009)Sutskever, Salakhutdinov, and
  Tenenbaum]{sutskever:nips09}
I.~Sutskever, R.~Salakhutdinov, and J.~Tenenbaum.
\newblock Modelling relational data using {B}ayesian clustered tensor
  factorization.
\newblock In \emph{Advances in Neural Information Processing Systems 22}, pages
  1821--1828. 2009.

\bibitem[Xiong et~al.(2010)Xiong, Chen, Huang, Schneider, and
  Carbonell]{xiong:sdm10}
L.~Xiong, X.~Chen, T.~Huang, J.~Schneider, and J.~Carbonell.
\newblock Temporal collaborative filtering with {B}ayesian probabilistic tensor
  factorization.
\newblock In \emph{SIAM International Conference on Data Mining (SDM)}, 2010.

\end{thebibliography}

\newpage

\appendix

\section{Supplementary Material}
\label{sec:appendix}
\begin{figure}[htpp]
\begin{center}
\includegraphics[scale=0.4]{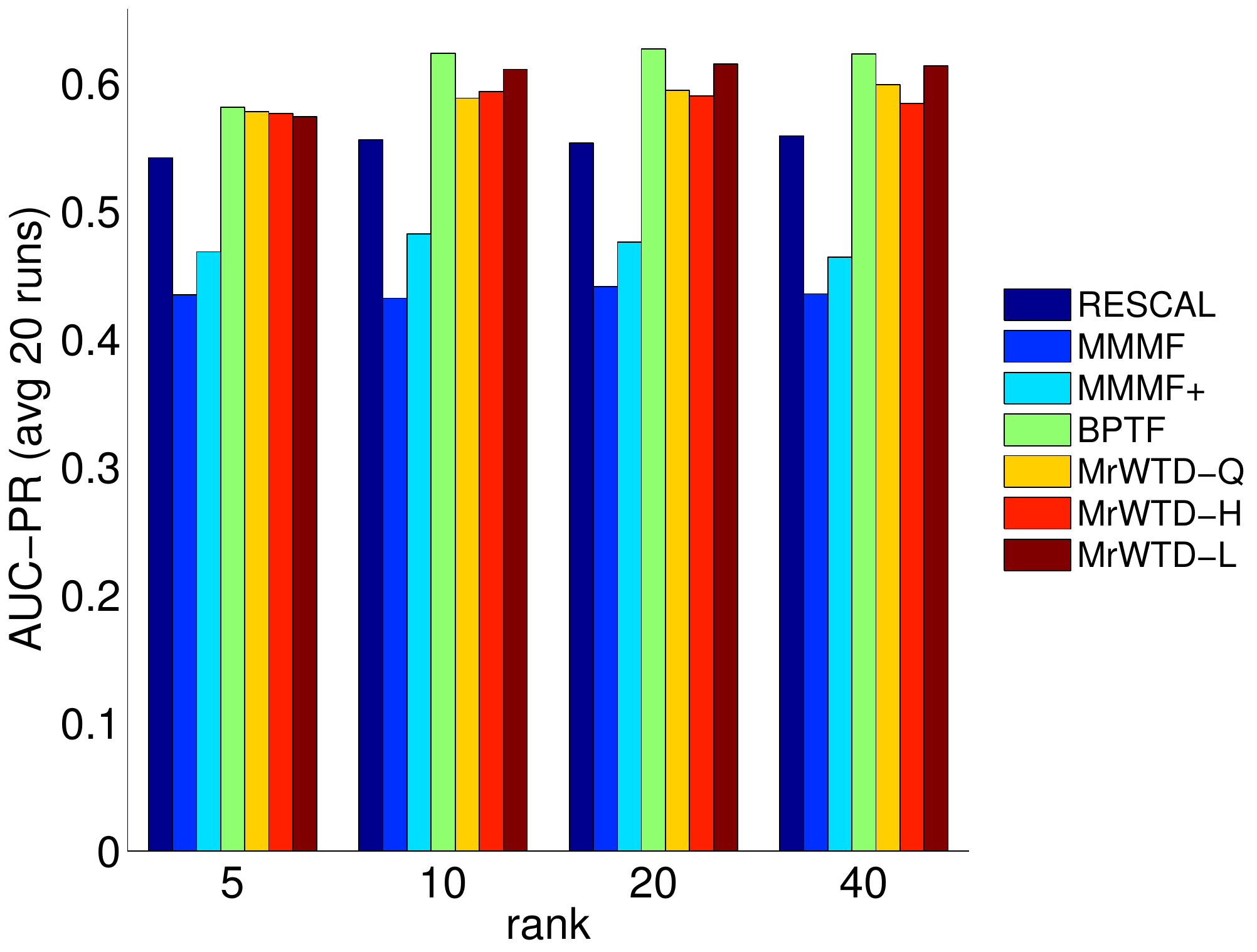}
\caption{Results of the rank data experiment (discussed in \autoref{sec:rank_exp}), using the Social Evolution dataset. The horizontal axis indicates the rank of the decomposition; the vertical axis indicates the area under the precision-recall curve, averaged over 20 runs. We use a random 25\% of the tensor for training data on each run. There is a slight increase from $\nrank=5$ to $\nrank=10$, but less than $0.02$ standard deviation (per algorithm) for $\nrank \geq 10$, supporting our hypothesis that regularization controls model complexity.}
\label{fig:rank_results}
\end{center}
\end{figure}

Here we report additional results from the experiments discussed in \autoref{sec:experiments}.
\autoref{fig:rank_results} shows the results of the rank experiment (\autoref{sec:rank_exp}).
\autoref{tab:timing} shows the results of the timing experiment (\autoref{sec:timing}).
In \autoref{tab:synth_results} and \autoref{tab:real_results}, we list the full results of the synthetic (\autoref{sec:synth_exp}) and real data (\autoref{sec:real_exp}) experiments.

\begin{table}[htdp]
\begin{center}
\caption{Results of the timing experiment (discussed in \autoref{sec:timing}). The  first column indicates the size of a synthetic dataset, measured by the number of objects $\nobj$; the remaining columns indicate the running time (in seconds), averaged over 10 runs, with the associated standard deviations. For each dataset, we use 10\% for training. Note that BPTF could not run on sizes $\{4000,8000\}$ due to runtime exceptions.}
\label{tab:timing}
\small{
\begin{tabular}{r | r r r r r r}
\hline
$\nobj$ & \multicolumn{2}{c}{RESCAL} & \multicolumn{2}{c}{BPTF} & \multicolumn{2}{c}{\Algo-H} \\
\hline
500 & 0.21 & (0.05) & 1.52 & (0.05) & 0.59 & (0.01) \\ 
1000 & 0.67 & (0.10) & 6.51 & (0.42) & 1.35 & (0.37) \\ 
2000 & 3.39 & (0.35) & 25.47 & (1.57) & 5.25 & (0.05) \\ 
4000 & 32.96 & (5.02) & -- & (--) & 20.33 & (0.80) \\ 
8000 & 332.77 & (40.88) & -- & (--) & 80.18 & (2.44) \\ 
\end{tabular}
}
\end{center}
\end{table}%

\begin{table*}[htbp]
\begin{center}
\caption{Results of the synthetic data experiments (discussed in \autoref{sec:synth_exp}). The first column indicates the dataset and score type. (For \SynthM, we provide two row groups to display the scores of  the binary- and real-valued slices.) The second column indicates the amount of training data (in percentage of the tensor). We evaluate on three variants of \Algo: -Q uses the quadratic loss for all slices; -H and -L use the quadratic loss for real-valued slices, but use the smooth hinge and logistic losses, respectively, for binary slices. For \SynthB and the first slice of \SynthM, we report area under the precision-recall curve (AUPRC); for the second slice of \SynthM, we report mean squared error (MSE). Standard deviations are listed in parentheses. All scores are averaged over 20 runs. Bold scores are statistically tied with the best score in each row.}
\label{tab:synth_results} 
\vspace{10pt}
\small{
\begin{tabular}{ l | l | l  l  l  l  l  l  l } 
\hline Data &Tr \%&RESCAL&MMMF&MMMF+&BPTF&MrWTD-Q&MrWTD-H&MrWTD-L\\ 
\hline
\multirow{6}{*}{\specialcell{Binary Synth \\ (AUPRC)}}
&3&0.14 (.00)&\textbf{0.13 (.04)}&0.13 (.00)&\textbf{0.15 (.01)}&0.13 (.01)&0.12 (.01)&0.13 (.01)\\ 
&5&0.16 (.01)&0.16 (.03)&0.17 (.01)&0.27 (.03)&0.23 (.03)&0.20 (.08)&\textbf{0.32 (.03)}\\ 
&10&0.21 (.03)&0.24 (.04)&0.36 (.01)&0.50 (.01)&0.55 (.02)&\textbf{0.70 (.02)}&\textbf{0.71 (.02)}\\ 
&15&0.38 (.03)&0.33 (.01)&0.51 (.01)&0.59 (.01)&0.66 (.02)&0.83 (.00)&\textbf{0.84 (.00)}\\ 
&20&0.49 (.02)&0.37 (.01)&0.61 (.01)&0.63 (.01)&0.70 (.02)&0.89 (.00)&\textbf{0.90 (.00)}\\ 
&25&0.57 (.02)&0.40 (.01)&0.69 (.01)&0.66 (.01)&0.73 (.02)&0.92 (.00)&\textbf{0.93 (.00)}\\ 
\hline
\multirow{6}{*}{\specialcell{Binary Synth \\ (AUPRC)}}
&3&0.14 (.01)&0.11 (.03)&0.13 (.01)&\textbf{0.26 (.02)}&\textbf{0.27 (.03)}&\textbf{0.27 (.03)}&\textbf{0.26 (.04)}\\ 
&5&0.18 (.01)&0.13 (.01)&0.18 (.03)&\textbf{0.37 (.02)}&\textbf{0.36 (.03)}&\textbf{0.37 (.03)}&\textbf{0.36 (.04)}\\ 
&10&0.27 (.02)&0.30 (.02)&0.35 (.03)&\textbf{0.57 (.02)}&0.55 (.03)&0.55 (.03)&0.54 (.04)\\ 
&15&0.33 (.01)&0.25 (.08)&0.48 (.01)&0.65 (.01)&\textbf{0.67 (.04)}&\textbf{0.67 (.02)}&\textbf{0.67 (.02)}\\ 
&20&0.39 (.02)&0.20 (.03)&0.57 (.02)&0.70 (.01)&\textbf{0.74 (.03)}&\textbf{0.74 (.02)}&\textbf{0.75 (.02)}\\ 
&25&0.43 (.01)&0.24 (.06)&0.66 (.01)&0.71 (.01)&\textbf{0.74 (.02)}&\textbf{0.74 (.02)}&\textbf{0.76 (.02)}\\ 
\hline
\multirow{6}{*}{\specialcell{Mixed Synth \\ (MSE)}}
&3&0.99 (.01)&0.99 (.01)&\textbf{0.43 (.08)}&0.48 (.03)&\textbf{0.44 (.04)}&\textbf{0.44 (.04)}&\textbf{0.45 (.03)}\\ 
&5&0.97 (.01)&0.96 (.02)&\textbf{0.12 (.07)}&0.24 (.02)&0.19 (.02)&0.19 (.01)&0.19 (.01)\\ 
&10&0.90 (.00)&0.92 (.01)&\textbf{0.03 (.01)}&0.06 (.01)&\textbf{0.05 (.04)}&0.04 (.00)&\textbf{0.05 (.04)}\\ 
&15&0.83 (.00)&0.90 (.02)&\textbf{0.02 (.00)}&0.03 (.00)&0.03 (.02)&0.03 (.02)&0.04 (.02)\\ 
&20&0.77 (.01)&0.84 (.00)&\textbf{0.02 (.01)}&\textbf{0.02 (.00)}&0.02 (.01)&0.03 (.01)&0.03 (.01)\\ 
&25&0.70 (.01)&0.81 (.00)&\textbf{0.01 (.00)}&0.01 (.00)&\textbf{0.01 (.00)}&\textbf{0.01 (.00)}&\textbf{0.01 (.00)}\\ 
\end{tabular} 
}
\end{center}
\end{table*}

\begin{table*}[htbp]
\centering
\caption{Results of the real data experiments (discussed in \autoref{sec:real_exp}). Standard deviations are listed in parentheses. All scores are averaged over 20 runs. Bold scores are statistically tied with the best score in each row. }
\label{tab:real_results} 
\vspace{10pt}
\small{
\begin{tabular}{ l | l | l  l  l  l  l  l  l } 
\hline Data &Tr \%&RESCAL&MMMF&MMMF+&BPTF&MrWTD-Q&MrWTD-H&MrWTD-L\\ 
\hline
\multirow{6}{*}{\specialcell{Kinship \\ (AUPRC)}}
&3&0.08 (.00)&0.08 (.00)&0.08 (.00)&\textbf{0.10 (.03)}&\textbf{0.10 (.04)}&0.09 (.01)&\textbf{0.11 (.03)}\\ 
&5&0.08 (.01)&0.09 (.00)&0.09 (.00)&0.26 (.03)&\textbf{0.33 (.02)}&0.31 (.04)&\textbf{0.33 (.02)}\\ 
&10&0.13 (.01)&0.11 (.00)&0.15 (.01)&\textbf{0.49 (.02)}&\textbf{0.44 (.13)}&0.46 (.03)&\textbf{0.48 (.02)}\\ 
&15&0.18 (.01)&0.14 (.01)&0.21 (.01)&0.57 (.01)&\textbf{0.61 (.01)}&0.42 (.24)&\textbf{0.57 (.12)}\\ 
&20&0.28 (.01)&0.16 (.01)&0.27 (.01)&0.59 (.01)&\textbf{0.65 (.02)}&\textbf{0.65 (.02)}&0.63 (.01)\\ 
&25&0.34 (.01)&0.18 (.00)&0.33 (.02)&0.61 (.01)&0.68 (.01)&\textbf{0.70 (.01)}&0.69 (.01)\\ 
\hline
\multirow{6}{*}{\specialcell{Reality \\ (AUPRC)}}
&3&0.09 (.01)&0.09 (.01)&0.09 (.01)&\textbf{0.11 (.01)}&0.09 (.01)&0.09 (.01)&0.09 (.01)\\ 
&5&0.09 (.01)&\textbf{0.12 (.08)}&0.10 (.01)&\textbf{0.13 (.02)}&0.11 (.02)&0.11 (.02)&0.11 (.02)\\ 
&10&0.10 (.01)&0.13 (.03)&0.13 (.02)&\textbf{0.20 (.03)}&0.17 (.03)&0.17 (.03)&\textbf{0.18 (.04)}\\ 
&15&0.11 (.01)&0.19 (.03)&0.17 (.03)&\textbf{0.24 (.03)}&\textbf{0.27 (.06)}&0.23 (.06)&\textbf{0.23 (.08)}\\ 
&20&0.13 (.01)&0.22 (.06)&0.21 (.04)&0.29 (.03)&\textbf{0.32 (.04)}&0.27 (.06)&\textbf{0.29 (.08)}\\ 
&25&0.14 (.01)&0.26 (.03)&0.25 (.05)&0.31 (.04)&\textbf{0.34 (.02)}&\textbf{0.31 (.05)}&\textbf{0.34 (.04)}\\ 
\hline
\multirow{6}{*}{\specialcell{Social \\ (AUPRC)}}
&3&0.34 (.03)&0.29 (.02)&0.30 (.01)&\textbf{0.41 (.01)}&0.32 (.01)&0.32 (.01)&0.32 (.01)\\ 
&5&0.35 (.03)&0.30 (.01)&0.31 (.01)&\textbf{0.45 (.02)}&0.35 (.03)&0.36 (.04)&0.36 (.03)\\ 
&10&0.36 (.01)&0.33 (.01)&0.36 (.01)&\textbf{0.51 (.01)}&0.50 (.02)&0.47 (.03)&0.43 (.06)\\ 
&15&0.45 (.01)&0.38 (.04)&0.41 (.02)&\textbf{0.56 (.01)}&0.54 (.02)&0.54 (.02)&0.54 (.03)\\ 
&20&0.52 (.02)&0.41 (.02)&0.44 (.02)&\textbf{0.60 (.02)}&0.57 (.01)&0.57 (.02)&0.58 (.02)\\ 
&25&0.56 (.01)&0.44 (.01)&0.47 (.01)&\textbf{0.63 (.01)}&0.60 (.01)&0.60 (.01)&0.62 (.01)\\ 
\end{tabular} 
}
\end{table*}

\end{document}